\documentclass[10pt,twocolumn,letterpaper]{article}

\usepackage{iccv}
\usepackage{times}
\usepackage{epsfig}
\usepackage{graphicx}
\usepackage{amsmath}
\usepackage{amssymb}

\usepackage{enumitem}
\usepackage{tabularx}
\usepackage{booktabs}
\usepackage{xcolor}
\usepackage[title]{appendix}

\renewenvironment{quote}
  {\list{}{\rightmargin=0.4cm \leftmargin=0.4cm}%
  \item\relax}
  {\endlist}

\definecolor{colform}{RGB}{56, 118, 28}
\definecolor{colcontent}{RGB}{230, 145, 56}
\definecolor{colcontext}{RGB}{17, 85, 204}

\usepackage[pagebackref=true,breaklinks=true,letterpaper=true,colorlinks,bookmarks=false]{hyperref}

\iccvfinalcopy 

\def\iccvPaperID{5743} 

\ificcvfinal\fi

\begin{document}

\title{Explain Me the Painting: Multi-Topic\\Knowledgeable Art Description Generation}

\author{Zechen Bai\thanks{Work done during a remote internship at Osaka University.}
\quad Yuta Nakashima$^{1}$
\quad Noa Garcia$^{1}$  \\
$^1$Osaka University  \\
{\tt\small zechenbai@outlook.com}
\quad{\tt\small n-yuta@ids.osaka-u.ac.jp}
\quad{\tt\small noagarcia@ids.osaka-u.ac.jp} \\
}

\maketitle
\ificcvfinal\thispagestyle{empty}\fi

\begin{abstract}
Have you ever looked at a painting and wondered what is the story behind it? This work presents a framework to bring art closer to people by generating comprehensive descriptions of fine-art paintings. Generating informative descriptions for artworks, however, is extremely challenging, as it requires to 1) describe multiple aspects of the image such as its style, content, or composition, and 2) provide background and contextual knowledge about the artist, their influences, or the historical period. To address these challenges, we introduce a multi-topic and knowledgeable art description framework, which modules the generated sentences according to three artistic topics and, additionally, enhances each description with external knowledge. The framework is validated through an exhaustive analysis, both quantitative and qualitative, as well as a comparative human evaluation, demonstrating outstanding results in terms of both topic diversity and information veracity.
\end{abstract}

\section{Introduction}

For the general public, art tends to be considered as a mysterious and remote discipline that requires a lot of study to be fully appreciated. In the last few years, many efforts have been made to apply artificial intelligence technologies to the domain of art to make it more accessible \cite{fiorucci2020machine}. Thanks to the large-scale digitisation of artworks from collections all over the world \cite{googlearts,wga,RijksmuseumWeb,metweb}, computer vision techniques have been widely adopted to address different art-related problems \cite{karayev2013recognizing,mensink2014rijksmuseum,crowley2014state,gatys2016image,mao2017deepart,lecoutre2017recognizing,chu2018image,strezoski2018omniart,Garcia2018How,wynen2018unsupervised}. 

Most of the existing work in the field is focused on the automatic analysis of paintings, addressing problems such as attribute prediction \cite{mensink2014rijksmuseum,strezoski2018omniart}, content analysis  \cite{crowley2014state,gonthier2018weakly} or style identification \cite{johnson2008image,shamir2010impressionism,karayev2013recognizing}. However, there is still an absence of research that conveys in-depth and comprehensive 
information of artworks for the general public. In other words, most previous work only allows people to understand a unique aspect of an artwork by providing a single tag, usually associated with either its style or its content. 

\begin{figure}
    \centering
    \includegraphics[width=0.99\linewidth]{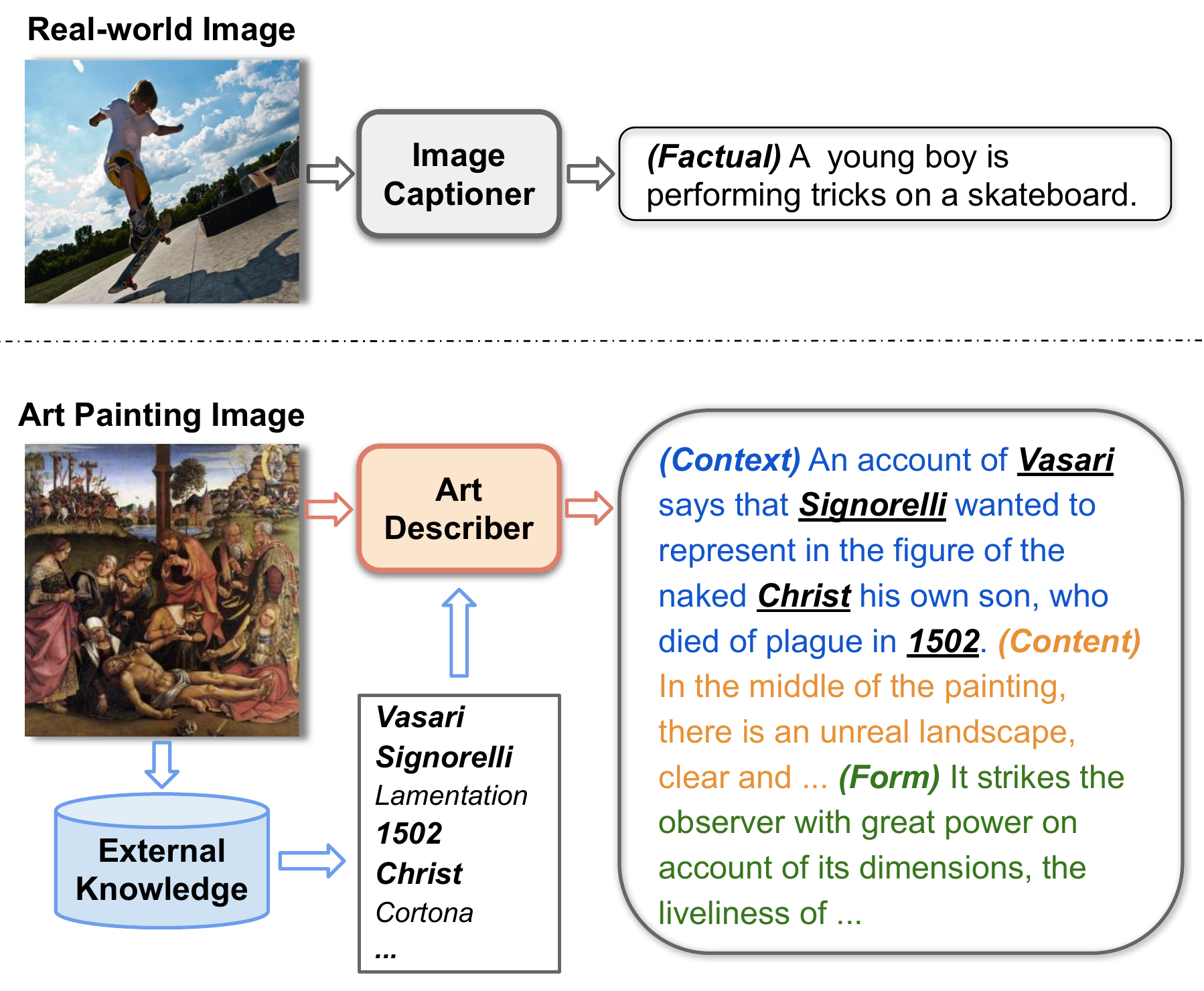}
    \caption{\textbf{Art description generation.} Comparison between standard image captioning (top) and art image description generation (bottom). Standard image captioning usually generates a single factual natural sentence to describe the content of a real-world image, while our art image description contains multiple sentences to explain an artwork from different artistic perspectives.}
    \label{fig:idea_illustration}
\end{figure}

However, a real understanding of art is much more complex than being able to successfully categorize each piece into a set of pre-defined tags. The intricate relationships between the \textcolor{colcontent}{\textbf{content}}, \textcolor{colform}{\textbf{form}}, and \textcolor{colcontext}{\textbf{context}} of each artwork are not at hand in a simple categorization process. In this work, we take a step forward on the field of art understanding and accessibility by proposing to automatically generate rich descriptions of paintings from multiple artistic perspectives. We propose a multi-topic and knowledgeable art description generation framework that produces detailed explanations about different aspects of paintings. Such a service would facilitate a deeper interaction between the general public and artworks, as well as potentially ease the work of art curators by automatically generating comments of paintings.

Art description generation may be seen as a problem similar to image captioning \cite{vinyals2015show,dai2017towards,anderson2018bottom}, which aims to naturally and factually describe the content of the scene in the image. Compared to conventional image captioning, however, generating descriptions for artworks faces two extra challenges. First, a comprehensive explanation of an artwork requires not only the factual description of its content, but also background knowledge, such as details about its author, the context of the creation process, and so on. This information is rarely contained in the artwork image itself. Second, according to art historians \cite{belton1996art}, an informative description of an artwork should address three main topics: \textcolor{colcontent}{\textbf{content}}, \textcolor{colform}{\textbf{form}}, and \textcolor{colcontext}{\textbf{context}}. While in standard image captioning only the \textcolor{colcontent}{\textbf{content}} is considered, the distinction among different artistic topics requires to handle a more complex language modeling scheme.

To address these challenges, we introduce a multi-topic and knowledgeable art description generation framework that: 1) introduces external knowledge into the description generation process, and 2) proposes a multi-topic language model to describe the different aspects of the painting. The main idea is illustrated in Fig.~\ref{fig:idea_illustration}, where an artwork image is used to both generate a description and retrieve related knowledge from an external source. Our framework follows three steps. Firstly, by training a language model, we generate a masked sentence with fillable slots for the concepts that require external information to be known, such as artist, date of creation, location, \etc. Moreover, the language model incorporates information about different artistic topics so that the masked sentences are generated according to each topic. Secondly, a knowledge retrieval module is employed to retrieve external information related to the painting from open access databases (\eg \textit{Wikipedia}\footnote{\url{https://www.wikipedia.org}}). Finally, we design a knowledge-filling module that extracts candidate words from the retrieved knowledge and selects the appropriate concepts for each slot. 

In our exhaustive experimental section, including quantitative comparisons, qualitative analysis, and human evaluation, we show that our framework generates satisfactory art descriptions more accurately and informatively than others. Overall, our main contributions are:
\begin{itemize}[noitemsep,topsep=0pt]
    \item We propose the first framework for art description generation that creates multi-topic long descriptions of fine-art paintings. So far, art description generation has been tackled as an image captioning task by only generating short factual sentences about artworks.
    \item We design a multi-topic language modeling module to generate multi-topic descriptions. Additionally, we annotate an art description dataset with sentence-topic labels based on art historians protocols \cite{belton1996art}, which we share publicly\footnote{\url{https://github.com/noagarcia/explain-paintings}} to inspire future work not only in art description but also in general art understanding.
    \item We leverage a knowledge retriever and train a knowledge filling module as a fill-in-the-blank task to incorporate art information relevant to each painting. This method can be easily applied to other domains.
\end{itemize}

\section{Related Work}

\paragraph{Artwork Analysis}
Computer vision techniques have been widely adopted to address art-related problems \cite{ikeuchi2008digitally,mensink2014rijksmuseum,wilber2017bam,shugrina2019creative}. A fundamental task in the field is to extract representative features that can capture the insights of the style \cite{karayev2013recognizing,gatys2016image,wynen2018unsupervised,huckle2020demographic} or the content of a painting \cite{crowley2014state,crowley2014search,seguin2016visual,gonthier2018weakly}  and use them for the automatic analysis of artworks, in tasks such as classification \cite{ma2017part,carneiro2012artistic,tan2016ceci,garcia2019context}, style identification \cite{johnson2008image,shamir2010impressionism,karayev2013recognizing,wynen2018unsupervised}, object recognition \cite{crowley2014state,crowley2015face,gonthier2018weakly}, or image retrieval \cite{carneiro2012artistic,crowley2014state,crowley2015face}.
Although art categorization is challenging due to the inherent diversity and abstraction in art, it only studies a single aspect of the artwork. However, paintings are complex images full of symbolism. A single label cannot totally represent the elaborated relationship between the depicted elements, the painter's motivations, and the historical context of the production. For a full comprehension of paintings, we propose to generate coherent language representations in the form of artistic descriptions. Until now, only a few studies \cite{sheng2016dataset,Garcia2018How,garcia2020dataset} have applied multimodal vision and language techniques to the domain of art. While in \cite{Garcia2018How} a system to find paintings given textual descriptions is proposed, other methods \cite{sheng2016dataset,garcia2020dataset} predicted answers to questions about artworks. However, generating comprehensive descriptions for fine-art paintings is still rarely studied.

\begin{figure*}[htb]
    \centering
    \includegraphics[width=0.99\linewidth]{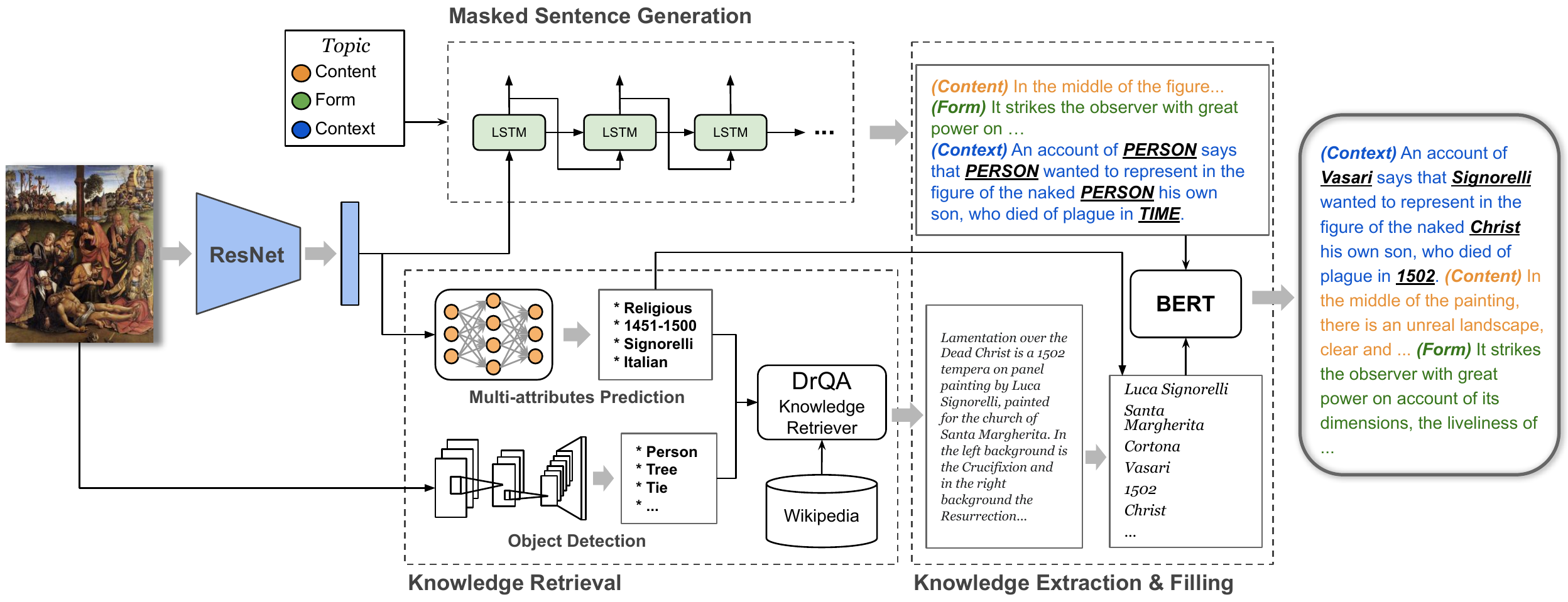}
    \caption{\textbf{Proposed framework.} It consists of three parts: masked sentence generation, knowledge retrieval, and knowledge filling.}
    \label{fig:framework}
\end{figure*}

\vspace{-12pt}
\paragraph{Image Captioning}
Encoder-decoder image captioning models for natural images are data-driven methods using deep neural networks \cite{vinyals2015show,xu2015show,anderson2018bottom,rennie2017self,yao2017incorporating,li2019pointing,zhou2019improving,lu2018neural,wang2020show}. The classic scheme \cite{vinyals2015show} combines a convolutional neural network (CNN) as image encoder and a recurrent neural network (RNN) as caption decoder. Several variations have emerged, such as adding attention \cite{xu2015show} or using detected objects instead of plain pixels \cite{anderson2018bottom}. Although these models obtain good results for natural images, they do not transfer well to cultural images \cite{sheng2019generating}. To generate descriptions for artworks, previous work introduced an ontology \cite{xu2017ontology} and a hierarchical model \cite{xu2015semantic}, leveraging low-level features, \eg, image texture and meta-data of cultural images, which heavily rely on feature engineering. Moreover, they could only generate a single and factual sentence about the image \textcolor{colcontent}{\textbf{content}}. In contrast, we generate multi-topic descriptions, relying on external sources to improve the information quality. External knowledge has been used in image captioning \cite{wu2017image,yao2017incorporating,li2019pointing,zhou2019improving,lu2018entity,biten2019good}, mostly by relying on available tags  \cite{lu2018entity} or texts \cite{biten2019good} associated to the target image. Differently, our external knowledge is retrieved by only using the image.

\section{Method}

Our framework contains three main parts: 1) masked sentence generation, 2) knowledge retrieval, and 3) knowledge extraction and filling. As shown in Fig.~\ref{fig:framework}, we first extract $D$-dimensional visual features from $L$ spatial locations from the image \cite{xu2015show} using a pre-trained ResNet \cite{he2016deep},\footnote{Despite the domain gap, ResNet pre-trained on ImageNet dataset has been shown to work well for art images \cite{sabatelli2018deep,strezoski2018omniart}.} $V=\{\mathbf{v}_1,\mathbf{v}_2,...,\mathbf{v}_L\},\mathbf{v}_i \in \mathbb{R}^D$. Then, in the masked sentence generation part, we input $V$ into the topic decoder to generate multi-topic masked sentences that describe the painting from multiple aspects. These masked sentences have blanked concepts to be filled with knowledge at a later stage. In the knowledge retrieval part, we take the average-pooling vector $\bar{\mathbf{v}} = \sum_i \mathbf{v}_i / L$ as the global visual feature for multi-attribute prediction. We also employ an object detector to detect visual concepts. The predicted attributes and the detected objects are used to retrieve relevant knowledge from an external source with DrQA \cite{chen2017reading}. Finally, in the knowledge extraction and filling part, given the generated multi-topic masked sentences and the retrieved knowledge text, we extract candidate knowledge concepts, and use a BERT-based model \cite{devlin2019bert} to get the final description.

\subsection{Masked Sentence Generation}
\label{sec:mask_sent_gen}
Traditional image captioning datasets, such as MSCOCO \cite{lin2014microsoft}, provide high-quality general captions. Image captioning decoders trained on these corpora predict a probability distribution over a closed vocabulary to generate text. However, these decoders have difficulties in generating specific entities that occur sparsely in the vocabulary. For example, in a vocabulary for art description, the artist name, location, or timeframe may all be in a low frequency. Moreover, the desired art descriptions should contain external knowledge not directly present in the image. We address these issues by only relying on the decoder to generate masked sentences, which are later completed by the knowledge extraction and filling module.

\subsubsection{Data Preprocessing} 
Given set $\mathcal{D}$ of descriptions about paintings, we obtain a training corpus for the masked sentence generation part by performing Named-Entity-Recognition (NER). Specifically, we apply Stanford CoreNLP name tagger \cite{manning2014stanford} to the descriptions to extract entities of the following types: {\small\texttt{person}}, {\small\texttt{location}}, {\small\texttt{organization}}, {\small\texttt{ordinal}}, {\small\texttt{number}}, {\small\texttt{date}}, and {\small\texttt{misc}}. Then, we replace the found entities in the description with their corresponding entity type, \eg:%
\begin{quote}
An account of \textbf{Vasari} says that \textbf{Signorelli} wanted to represent in the figure of the naked \textbf{Christ} his own son, who died of plague in \textbf{1502}.
\end{quote} %
is transformed into %
\begin{quote}An account of {\small\texttt{[person]}} says that {\small\texttt{[person]}} wanted to represent in the figure of the naked {\small\texttt{[person]}} his own son, who died of plague in {\small\texttt{[date]}}. 
\end{quote}

\subsubsection{Topic Decoder}
\label{sec:topicdecoder}
We envisage a decoder that can handle multi-topic description generation: given a visual feature $V$ and a desired topic $d$, the decoder should generate corresponding topic-related masked sentences. In this section, we first introduce a baseline decoder that generates topic-agnostic masked sentences. Then, we explore two variants to address the multi-topic challenge. Finally, we explain how the multi-topic masked description is generated. Figures illustrating each decoder can be found in Appendix A.

\vspace{-10pt}
\paragraph{Baseline Decoder} Following Xu \etal \cite{xu2015show}, we employ a long short-term memory (LSTM) \cite{hochreiter1997long}-based decoder to decode image visual features into masked sentences. The decoder generates one word $\mathbf{y}_t$ (in the one-hot vector representation) at each time step $t$ based on the attention-based visual context vector $\mathbf{z}_t$, the previous hidden state $\mathbf{h}_{t-1}$, and the previously generated word $\mathbf{y}_{t-1}$. Formally:
\begin{align}
\label{eq:lstm}
    \mathbf{h}_t &= \text{LSTM}([\mathbf{z}_t, \mathbf{h}_{t-1}, \mathbf{E}{\mathbf{y}_{t-1}}]) \\
\label{eq:f_att}
    g_{ti}&=f_{att}(\mathbf{v}_i, \mathbf{h}_{t-1}) \\
\label{eq:att_softmax}
    \boldsymbol{\alpha}_{t}&=\text{softmax}(\mathbf{g}_{t}) \\
\label{eq:att_sum}
    \mathbf{z}_t&=\sum_{i=1}^{L}\mathbf{\alpha}_{ti}\mathbf{v}_{i}, 
\end{align}
where $[\cdot]$ denotes concatenation, $\mathbf{E}$ is an embedding matrix, $f_{att}$ is a trainable function for predicting the attention weights, for which we use a multilayer perceptron, and $\boldsymbol{\alpha}_{t}=\{\alpha_{t1},\dots,\alpha_{tL}\}$ are attention scores that sum to one. The visual context vector $\mathbf{z}_t$ is a dynamic representation of the relevant part of the input image at time $t$. Based on the LSTM state $\mathbf{h}_t$ and visual context vector $\mathbf{z}_t$, we calculate the output word probability with a fully-connected layer as:
\begin{equation}
    p(\mathbf{y}_t|\mathbf{y}_{1:t-1}, V)=\text{softmax}(\mathbf{W}_y[\mathbf{h}_t,\mathbf{z}_t]+\mathbf{b}_y),
\end{equation}
where $\mathbf{W}_y$ and $\mathbf{b}_y$ are the parameters in the fully-connected layer. Given the ground truth masked sequence $\mathbf{y}_{1:T}^{*}$, the weights of the decoder are optimized by minimizing the negative log-likelihood in training as:
\begin{equation}
    \mathcal{L}_\text{mle}=-\sum_{t=1}^T \log p(\mathbf{y}_t^* | \mathbf{y}_{1:t-1}^*,V).
\end{equation}

\vspace{-10pt}
\paragraph{Topic Parallel Decoder}
Due to the linguistic distinctions that different artistic topics may present, we propose to use different decoders to generate masked sentences for different topics independently, \ie, using different baseline decoders as \textit{sub-decoders} for each topic, respectively. This parallel setting is intuitive as it divides topic-related sentences into different decoding branches, enabling the decoders to not disturb each other. Formally, the parallel decoder can be formulated as:
\begin{equation}
    \mathbf{y}_{1:T}^{(d)}=\text{Parallel}(V, d)
\end{equation}
where $d$ is the topic label, which also serves as a selector for the sub-decoders. Within each sub-decoder, the computation is the same as in the baseline decoder:
\begin{equation}
    \mathbf{h}_t^{(d)} = \text{LSTM}^{(d)}([\mathbf{z}_t^{(d)}, \mathbf{h}_{t-1}^{(d)}, \mathbf{E}^{(d)} \mathbf{y}_{t-1}^{(d)})
\end{equation}
Equations (\ref{eq:f_att})--(\ref{eq:att_sum}) can be written equivalently. The different sub-decoders are optimized separately during training. 

\vspace{-10pt}
\paragraph{Topic Conditional Decoder}
To improve the computational efficiency as well as to leverage common knowledge among the different topics, we also propose a single-model solution for multi-topic description generation. Inspired by stylized captioning \cite{guo2019mscap}, we explore a topic conditional decoder. The conditional decoder injects a topic conditional vector into the decoding process, formulated as:
\begin{equation}
    \mathbf{y}_{1:T}^{(d)}=\text{Conditional}(V, d)
\end{equation}
Specifically, the topic label $d$ is transformed into a $N_\text{topic}$-dimensional one-hot vector $\mathbf{d}'$ to represent $N_\text{topic}$ topics, where each element represents the corresponding topic. Then, we feed $\mathbf{d}'$ into a topic embedding layer and concatenate the resulting vector with the standard inputs of the baseline decoder as:
\begin{equation}
    \mathbf{h}_t = \text{LSTM}([\mathbf{z}_t, \mathbf{h}_{t-1}, \mathbf{E} \mathbf{y}_{t-1},\mathbf{E}_\text{topic} \mathbf{d}'])
\end{equation}
where $\mathbf{E}_\text{topic}$ is the topic embedding matrix. To ensure that the generated masked sentences correctly contain the target topic, we employ a topic classifier $TC$ for constrain, \ie, $TC(\text{Conditional}(V, d)) \xrightarrow{} d$. The topic classifier is implemented as in TextCNN \cite{kim2014convolutional} and is jointly optimized with the topic conditional decoder with a classical \textit{cross-entropy loss} $\mathcal{L}_\text{ce}$. Overall, the objective $\mathcal{L}_\text{cond}$ of topic conditional decoder is:
\begin{equation}
    \mathcal{L}_\text{cond}=\mathcal{L}_\text{mle}+\mathcal{L}_\text{ce}
\end{equation}

\paragraph{Multi-Topic Masked Description} Given $N_\text{topic}$ topics, we generate $N_\text{topic}$ masked sentences, one per topic, with either the parallel or the conditional topic decoder. The full multi-topic masked description is then the concatenation of the $N_\text{topic}$ independently generated masked sentences.

\subsection{Knowledge Retrieval}
\label{sec:knowledge_retrieval}
To address the challenge of generating informative descriptions, we rely on external knowledge bases, such as Wikipedia. We use DrQA \cite{chen2017reading}, an efficient document retriever, to find relevant information. Formally, a given query text $q$ and all articles $c_j \in C$ in the knowledge base are tokenized and encoded as TF-IDF vectors, denoted as $\hat{\mathbf{q}}$ and $\hat{\mathbf{c}}_j$, respectively. Then, a similarity score is computed as: %
\begin{equation}s_j= \frac{\hat{\mathbf{q}}^\top \hat{\mathbf{c}}_j}{\|\hat{\mathbf{q}}\|\;\|\hat{\mathbf{c}}_j\|}
\end{equation}

Unlike previous work that uses off-the-shelf tags \cite{lu2018entity} or text \cite{biten2019good} as query to the knowledge base, we automatically constitute a query from the image by extracting: 1) \textit{Artistic attributes} using a multi-task attribute prediction model \cite{garcia2019context}. Specifically, for each painting we predict its {\small\texttt{artist}}, {\small\texttt{type}}, {\small\texttt{timeframe}}, and {\small\texttt{school}}; and 2) \textit{Visual concepts} using an object detection model \cite{ren2015faster} pre-trained in Visual Genome \cite{krishna2017visual}.
Concepts such as {\small\texttt{person}}, {\small\texttt{apple}}, \etc are extracted to describe the general content of the image, removing visual concepts that are unlikely to appear in paintings, such as {\small\texttt{cell phone}}. The words from the two sources are appended together to constitute our $q$. As an output of the knowledge retrieval module, we return the top-5 article $c_j$'s with the highest score $s_j$'s. To improve the ranking accuracy, we pre-process each $c_j$ as 1) stop word removal and word stemming, and 2) bigram TF-IDF. 

\subsection{Knowledge Extraction and Filling}
\label{sec:knowledge_fill}
In this module, we fill the masked concepts in the generated multi-topic masked descriptions with one or several knowledge words. Given the top-5 articles from the knowledge retrieval part, we further narrow down the knowledge space by extracting named entities, again using Stanford CoreNLP. We use the extracted named entities and the \textit{artistic attributes} from Sec.~\ref{sec:knowledge_retrieval} to compose a set of candidate words $G$. Then, we train a BERT-based model \cite{devlin2019bert} as a sequence to sequence task to find appropriate words from $G$ to fill the blanks in the generated multi-topic masked descriptions. Specifically, we generate an input sequence as:
\begin{equation}
SEQ_\text{in} = [\texttt{[CLS]}, y, \texttt{[SEP]}, k]
\end{equation}
where $k$ is a sequence with the concatenation of all the words in $G$ and $y$ denotes the sequence of words in the multi-topic masked description. The output description is:
\begin{equation}
SEQ_\text{out} = \text{BERT}(SEQ_\text{in})
\end{equation}
where $\text{BERT}(\cdot)$ is trained by minimizing the \textit{cross-entropy loss} to produce the original image descriptions in $\mathcal{D}$.

\section{Experiments}
Here we describe the experiments and their results. Implementation details can be found in Appendix B.

\vspace{-10pt}
\paragraph{Art Dataset} We use the SemArt dataset \cite{Garcia2018How}, which consists on $21,384$ painting images. Each image is associated with an artistic comment and seven attributes, such as {\small\texttt{artist}}, {\small\texttt{title}}, or {\small\texttt{date}}. The dataset is split into $19,244$ images for training, $1,069$ for validation, and $1,069$ for test.

\vspace{-12pt}
\paragraph{Artistic Topic Annotation} To investigate multi-topic description generation, we annotate the original comments in SemArt with their correspondent artistic topic. Following art historian protocols \cite{belton1996art} we use three topics: 1) \textcolor{colcontent}{\textbf{content}}, which describes what the artwork is about, \ie the \textit{message}; 2) \textcolor{colform}{\textbf{form}}, which describes how the work looks, \ie, the constituent elements of the work independent of their meaning; and 3) \textcolor{colcontext}{\textbf{context}}, which describes in what circumstances the work is or was. We rely on Amazon Mechanical Turk\footnote{\url{https://www.mturk.com/}} (AMT). We split the original comments on the train and test set into individual sentences and ask workers to annotate each sentence with one of the three topics. Workers are exposed to the image, the original full comment, the title, the artist name, and the year of creation. In total, $17,249$ images along with $33,543$ sentences are annotated.\footnote{We exclude some images with non-meaningful coments from the annotation, \eg ``Catalogue numbers: F 526".}
\vspace{-10pt}

\paragraph{Knowledge Base}
As external source of information, we use the 2016-12-21 dump\footnote{\url{https://dumps.wikimedia.org/enwiki/latest/}} of English Wikipedia. For each page, only the plain text is extracted. All structured and non-text data sections such as lists and figures are stripped. After discarding internal disambiguation, list, index, and outline pages, we retain $5,075,182$ articles.

\subsection{Human Evaluation}
\label{sec:humaneval}
The evaluation of generated text is a challenging task~\cite{celikyilmaz2020evaluation}, due to the complexity of automatically measuring not only grammatical correctness but also veracity, informativeness, and diversity. Automatic metrics designed to evaluate factual tasks such as machine translation (\eg, BLEU~\cite{papineni2002bleu}) or image captioning (\eg, CIDEr~\cite{vedantam2015cider}) do not work well on more creative tasks such as ours. Following previous work~\cite{fan2018hierarchical,salvador2019inverse,toivanen2012corpus}, we based our evaluation on how well humans perceive the text generate by our models.

We conduct a human evaluation on AMT on 100 randomly selected validation paintings. For each painting, we show a generated description to 3 annotators, together with the image, the original SemArt comment, the title, the artist, and the creation year. We ask annotators to rate each description according to the metrics below (higher is better):

\begin{table*}[t]
\small
\setlength{\tabcolsep}{5pt}
\centering
\caption{\textbf{Human evaluation.} Human ratings (mean and standard deviation) on generated descriptions according to six metrics.}
\begin{tabularx}{0.99\textwidth}{l l c c c c c c}
\toprule
Model & Knowledge & Understand & Relevance & Veracity & Content & Form & Context \\
\midrule
SAT \cite{xu2015show} (Baseline) & - &	$\textbf{3.62}\pm0.63$ & $1.94\pm0.94$ & $1.30\pm0.70$ & $0.35\pm0.48$ &	$0.05\pm0.21$ &	$0.65\pm0.48$ \\
Ours (Wikipedia) & Retrieved Wikipedia & $2.71\pm0.64$ & $2.29\pm1.04$ & $1.56\pm0.64$ & $0.73\pm0.45$ & $0.33\pm0.47$ & $0.83\pm0.38$ \\
Ours (SemArt) & Retrieved SemArt &	$2.77\pm0.62$ & $2.02\pm1.08$ & $1.39\pm0.59$ & $0.75\pm0.44$ & $0.37\pm0.48$ & $0.90\pm0.30$ \\
Ours (Oracle) & Original SemArt &	$2.71\pm0.64$ & $\textbf{2.49}\pm1.00$ & $\textbf{1.72}\pm0.70$ & $\textbf{0.76}\pm0.43$ & $\textbf{0.38}\pm0.49$ & $\textbf{0.91}\pm0.29$ \\
\bottomrule
\end{tabularx}
\label{tab:humanevaluation}
\end{table*}

\begin{figure*}[htb]
    \centering
    \includegraphics[width=0.97\linewidth]{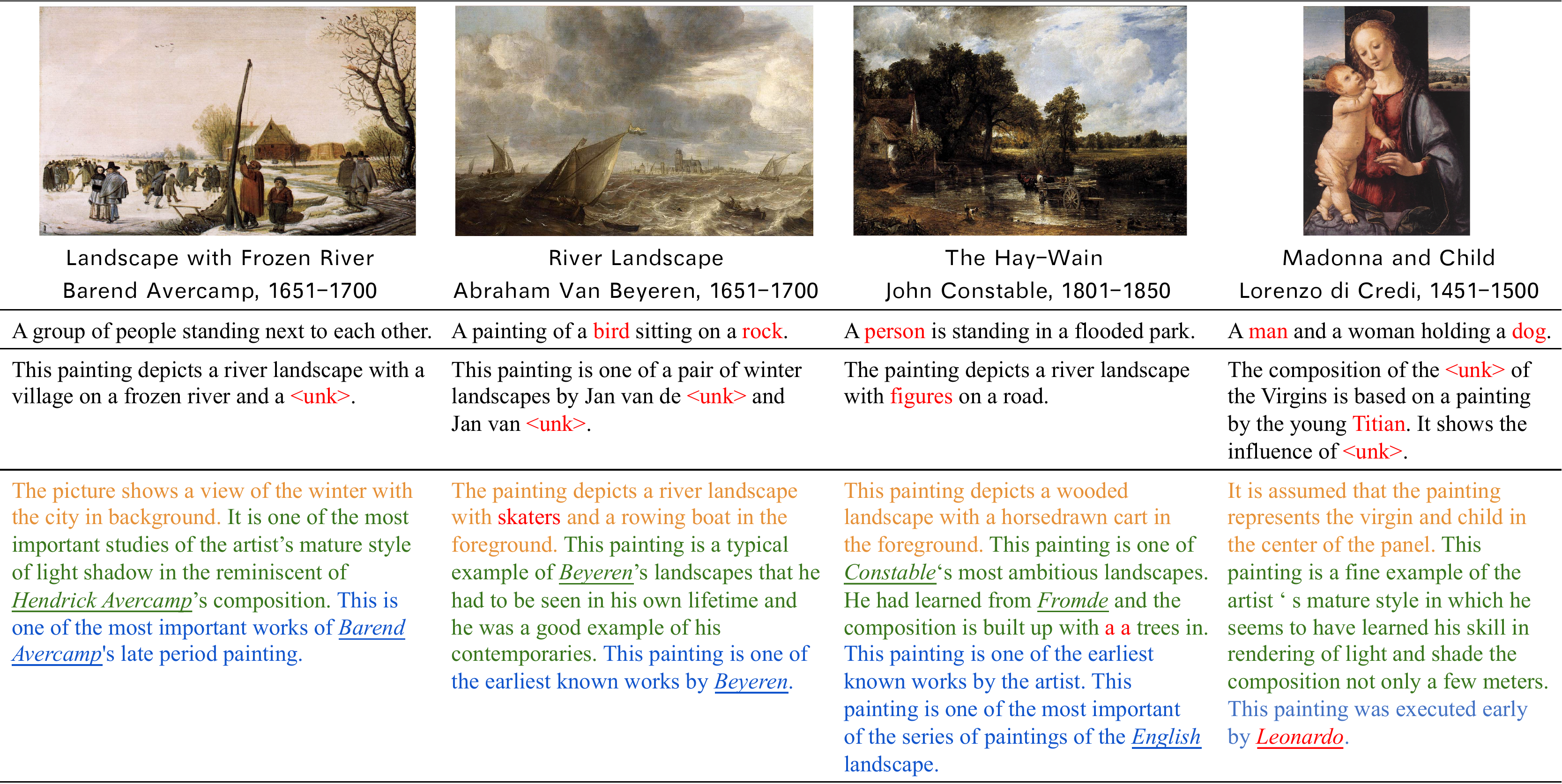}
    \caption{\textbf{Qualitative evaluation.}  Top row shows four test paintings together with their title, artist, and creation timeframe. The next rows contain the descriptions generated by SAT-Transfer, SAT-Baseline, and our method, respectively. Incorrect words are highlighted in \textcolor{red}{red}.}
    \label{fig:quality_examples}
\end{figure*}

\begin{itemize}[noitemsep,topsep=0pt]
\item \textit{Understandable}: integer from 1 to 4 measuring if the description can be understood by a human.

\item \textit{Relevance}: integer from 1 to 4 measuring if the description is relevant to the given painting. 

\item \textit{Veracity}:  integer from 1 to 4 measuring if the description is correct according to the given information. 

\item \textit{Content existence}: 1 if the description contains information about the topic \textcolor{colcontent}{\textbf{content}}, and 0 otherwise.

\item \textit{Form existence}: 1 if the description contains information about the topic \textcolor{colform}{\textbf{form}},  and 0 otherwise.

\item \textit{Context existence}: 1 if the description contains information about the topic \textcolor{colcontext}{\textbf{context}},  and 0 otherwise.
\end{itemize}

Results for different variants of our proposed framework are summarized in Table~\ref{tab:humanevaluation}. The baseline model (SAT~\cite{xu2015show}) do not use topic modeling or external knowledge. The other three models (denoted as Ours) use the parallel decoder, the knowledge retrieval, and knowledge extraction and filling modules. The main difference between them lies in the source of external knowledge used at inference. Specifically,  in Ours (Wikipedia), we use Wikipedia as knowledge source. In Ours (SemArt), we build a knowledge base with the comments in SemArt and find the most relevant one with knowledge retrieval module. Finally, in Ours (Oracle), we use the original associated comments, assuming a perfect accuracy in the knowledge retrieval module.

Results show that the baseline achieves the highest score in the \textit{Understandable} metric. This is because the text generated is simpler and shorter than the text generated with our models: \eg, the average number of words of SAT output is 30.9, while in Ours (Oracle) it is 71.8. The shorter the sentence, the less prone to contain grammatical errors. Nevertheless, from all the other metrics, which are more related to the informative aspect of the description, we observe that our framework outperforms the baseline by a large margin. 

When comparing our three different settings, Ours (Oracle) achieves the best performance in all the metrics, with a large margin in \textit{Relevance} and \textit{Veracity}. This is natural, as the ground truth comments are used as the source of knowledge. In the existence metrics (\ie \textit{Content}, \textit{Form}, and \textit{Context}), the performance of the three models is very close, all of them outperforming the baseline by a large margin. This shows the effectiveness of the topic decoder, which is able to produce different types of sentences. Ours (Wikipedia) performs the worst among the three settings, as it uses the most challenging source of knowledge and also suffers from a domain-gap between the source of training (SemArt comments) and the source of testing (Wikipedia articles).

\subsection{Qualitative Analysis and Examples}
We further explore the results of our framework with a qualitative analysis in Fig.~\ref{fig:quality_examples}. We compare three methods. In the first row, we use SAT \cite{xu2015show} trained on MSCOCO, namely SAT-Transfer. The generated sentences are very similar to the standard captions generated for natural images. This model  suffers from: 1) not containing art-specific information, and 2) a visual domain-gap when transfering from natural images to paintings, with  miss-detected concepts, such as `dog' and `person', in \textit{Madonna and Child} and \textit{The Hay-Wain}, respectively. In the second row, SAT is trained on the SemArt dataset, \ie SAT-Baseline. The results show that it can generate some short and understandable sentences as in \textit{The Hay-Wain}. The main problem, however, is that it does not contain background knowledge, leaving some specific knowledge words as {\small\texttt{unk}} in the output. Finally, the last row shows the proposed framework with the parallel decoder and SemArt as source of knowledge. The topic decoder effectively generates different sentences for each topic, shown in different colors. Moreover, each sentence includes relevant knowledge, such as `Barend Avercamp' in \textit{Landscape with Frozen River}, and `Beyeren' in \textit{River Landscape}. More examples can be found in Appendix E.

These results also reveal some of our method limitations. Missdetection of the visual content (a common mistake in traditional image captioning) is shown in \textit{River Landscape}, where the boats are confused by `skaters'. In \textit{The Hay-Wain}, there is a syntax error (`a a') in the \textcolor{colform}{\textbf{form}} sentence, which is produced in the knowledge filling part, based on BERT, \ie it does not appear in the masked-sentence decoder based on LSTM. We hypothesized that it is because the size of training set is relatively small for BERT to learn the language structure of our task. This can be solved by applying language augmentation techniques. Finally, in \textit{Madonna and Child}, there is an imperfect-knowledge mistake, in which the painting is assigned to the wrong artist.

\subsection{Modules Analysis}
\begin{figure}[t]
    \centering
    \includegraphics[width=0.95\linewidth]{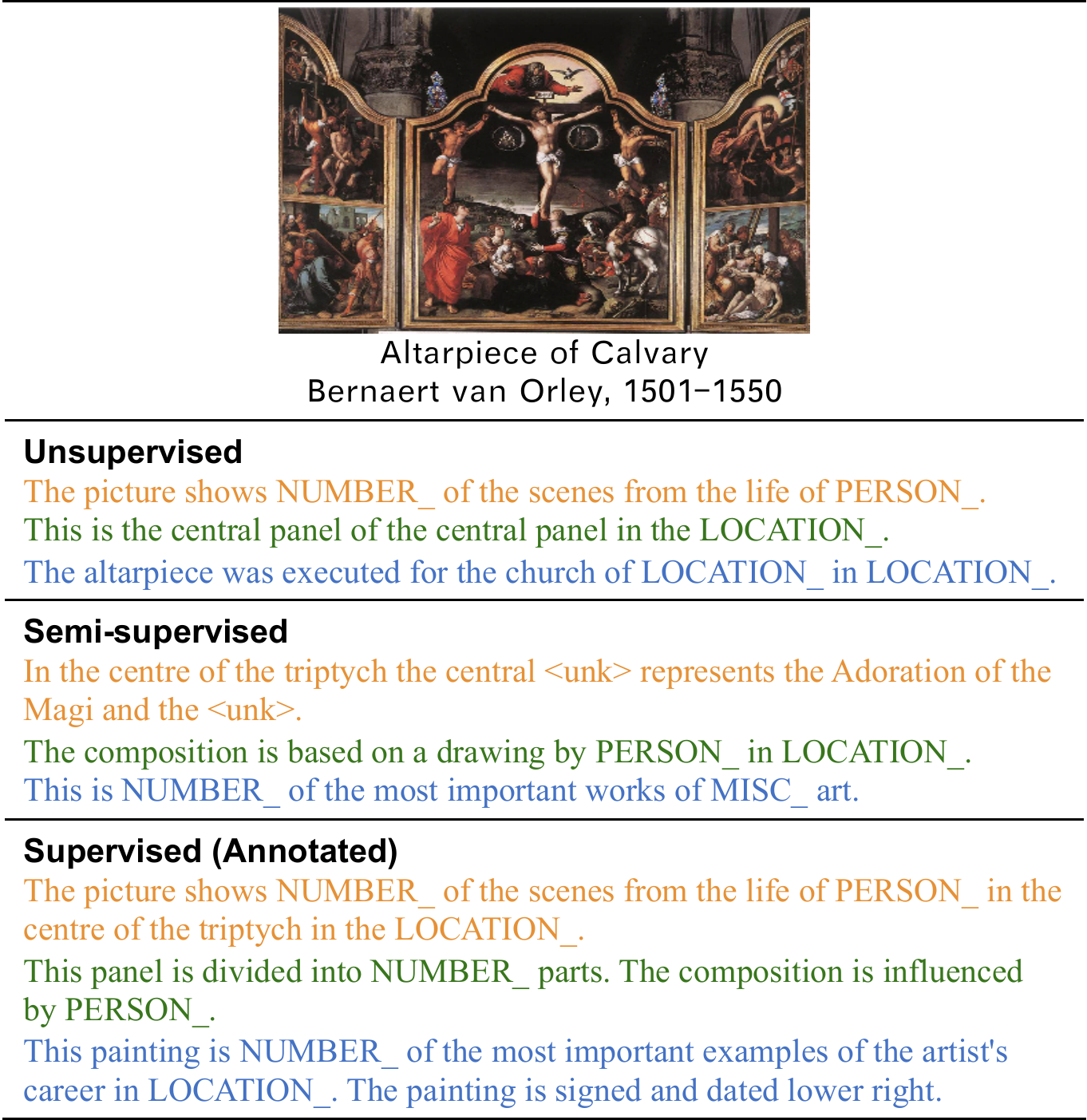}
    \caption{\textbf{Topic model comparison.} The topic labels are predicted with different settings to train the topic decoder.} 
    \label{fig:topic_example}
\end{figure}

\paragraph{Exploring Artistic Topic} Our topic decoder relies on the availability of a corpus with topic labels for training. For a more general approach, we explore un/semi-supervised topic models to automatically predict the topic label for each sentence. For the unsupervised setting, we use Latent Dirichlet Allocation (LDA) \cite{blei2003latent}, which assumes each topic is a mixture over an underlying set of words. While for the semi-supervised setting, we use Guided-LDA \cite{jagarlamudi-etal-2012-incorporating}, which incorporates lexical priors, manually set as a list of \textit{seed} words, to LDA. In our case, the \textit{seed} words are obtained from a subset of sentences with topic labels, from which we select the top-10 words with the highest frequency. The larger the subset, the more accurate the \textit{seed} words obtained. Here we use 3245 samples. The artistic comments are split into individual sentences and pre-processed to optimize each topic model method. After training, the topic model predicts a pseudo topic label for each sentence and used to train our parallel topic decoder.

LDA achieves $43.3$\% accuracy when predicting topic labels on the test set, while Guided-LDA achieves $51.6$\%. To reduce the influence of the knowledge filling module, in Fig.~\ref{fig:topic_example} we compare the generated masked sentences for the different topic models. In the unsupervised setting, the \textcolor{colcontent}{\textbf{content}} and \textcolor{colcontext}{\textbf{context}} sentences are generated correctly, while the \textcolor{colform}{\textbf{form}} sentence is confusing. In the semi-supervised setting, the output looks better, even close to the supervised approach, revealing that only a small amount of topic annotations may be necessary for art description generation.

\vspace{-15pt}
\paragraph{Knowledge Retrieval} The impact of the knowledge retrieval module and how it affects the overall framework has been already discussed in Sec.~\ref{sec:humaneval}. According to Table~\ref{tab:humanevaluation}, the easiest to retrieve the knowledge, the highest the scores. In Appendix D, we provide accuracy rates for the knowledge retrieval module on its own with different settings. It can be seen that the performance of the knowledge retrieval module is a crucial bottleneck on our system.

\begin{table*}[t]
\small
\centering
\caption{\textbf{Comparative evaluation.} We compare our models against multiple alternatives. All the methods are trained on the SemArt dataset under equivalent conditions. Methods using external knowledge (LSA and Ours) use Wikipedia as knowledge source. GM:GreedyMatching, S-T: Skip-Thought, EA: EmbeddingAverage.}
\begin{tabularx}{\textwidth}{r l >{\raggedleft\arraybackslash}p{1.4cm} >{\raggedleft\arraybackslash}p{1.4cm} >{\raggedleft\arraybackslash}p{1.4cm} >{\raggedleft\arraybackslash}p{1.4cm} >{\raggedleft\arraybackslash}p{1.4cm} >{\raggedleft\arraybackslash}p{1.4cm} >{\raggedleft\arraybackslash}p{1.4cm}}
\toprule
& Model & BLEU-4 & CIDEr & METEOR & ROUGE-L & GM & S-T & EA \\
\midrule
1 & NIC \cite{vinyals2015show} & 7.3 & \textbf{39.4} & 10.9 & \textbf{28.6} & 71.5 & 24.8 & 64.5 \\
2 & SAT \cite{xu2015show} (Baseline) & 6.5 & 38.6 & 11.1 & 27.5 & 72.9 & 26.6 & 73.3 \\
3 & Att2in \cite{rennie2017self} & 4.2 & 26.4 & 9.7 & 25.4 & 69.1 & 22.4 & 59.1 \\
4 & MScap \cite{guo2019mscap} & 0.4 & 0.1 & 6.3 & 14.2 & 64.0 & 19.3 & 50.7 \\
5 & OSCAR \cite{li2020oscar} & 0.1 & 2.0 & 2.8 & 11.3 & 63.0 & 33.3 & 84.8 \\
6 & LSA \cite{steinberger2004using} & 0.2 & 0.1 & 8.5 & 10.9 & 75.1 & \textbf{37.4} & 90.6 \\
7 & Ours (Parallel decoder) & \textbf{8.8} & 9.1 & \textbf{11.4} & 23.1 & \textbf{77.6} & 30.9 & \textbf{92.6} \\
8 & Ours (Conditional decoder) & 0.9 & 0.4 & 5.8 & 14.8 & 70.1 & 27.7 & 89.3 \\
\bottomrule
\end{tabularx}
\label{tab:sota_compare}
\end{table*}

\subsection{Comparative Evaluation}
\label{sec:compare_evaluation}
Although automatic metrics may not correlate well with the evaluation of the knowledge and creativity required to describe art, we include a standard automatic evaluation for completeness. We compare our proposed model against:

\vspace{-12pt}
\paragraph{Classic image captioning} (1)~NIC~\cite{vinyals2015show}, LSTM-based encoder-decoder model without attention; (2)~SAT~\cite{xu2015show}, which incorporates soft-attention (note that this corresponds to the baseline decoder in Sec.~\ref{sec:topicdecoder}); and (3)~Att2in~\cite{rennie2017self}, similar to SAT but in where the attention-derived context visual feature is only input to the cell node of the LSTM.

\vspace{-12pt}
\paragraph{Stylized image captioning} Our conditional topic decoder can be seen as a use case of stylized image captioning. We compare our framework with (4)~MScap~\cite{guo2019mscap}. In our re-implementation, we regard  \textit{topic} as \textit{style}.

\vspace{-12pt}
\paragraph{Transformer-based image captioning} We evaluate the state-of-the-art Transformer (5)~OSCAR~\cite{li2020oscar}, which is one of the latest multi-modal approaches in vision-language task. We use its original pre-trained weights and finetune it on the SemArt dataset.

\vspace{-12pt}
\paragraph{Text-summarization} As a different perspective, we generate painting descriptions with text-based summarization methods. That is, given the knowledge articles retrieved in Sec.~\ref{sec:knowledge_retrieval}, we summarize their content with (6) LSA \cite{steinberger2004using}, a language-independent algebraic method.

\vspace{-12pt}
\paragraph{Ours} We compare our method when using the (7) Topic Parallel Decoder and the (8) Topic Conditional Decoder.

Table~\ref{tab:sota_compare} shows the comparison results. We adopt a wide range of natural language evaluation metrics, including BLEU-4 \cite{papineni2002bleu}, CIDEr \cite{vedantam2015cider}, METEOR \cite{denkowski2014meteor}, ROUGE-L \cite{lin2004rouge}, GreedyMatching~\cite{rus2012optimal}, Skip-Thought~\cite{kiros2015skip}, and EmbeddingAverage~\cite{liu2016not}. We notice that (1) the scores are much lower than in traditional image captioning: \eg, the BLEU-4 score in MSCOCO is around 30 \cite{vinyals2015show,xu2015show}, while in SemArt it is less than 10; (2) the scores among different metrics shows very large variation. On one hand, this phenomenon shows that the proposed task is very challenging. On the other hand,  
these automatic metrics may not be appropriate to evaluate the richness and diversity of art description generation. Thus we also include the qualitative comparison results in Appendix E.

Firstly, our parallel decoder performs better than the conditional decoder. It is natural as the parallel decoder has larger capacity in network and can fit different topics in a specific manner. When comparing against the classic image captioning (NIC, SAT, and Att2in), our parallel decoder method outperforms them in 5 out of 7 metrics. The performance gain is mainly caused by the masked-sentence-generation-filling schema that reduces the burden of the decoder to handle low frequency words and named-entities. In MScap, we directly let the model to conditionally learn the topic, without the help of masked-sentence-generation-filling, the performance even getting worse. With respect to the Transformer-based approach OSCAR, it performs very poorly in most of the metrics. We deduce this is caused by: 1) the scale of the training set, and 2) the use of Bottom-Up-Top-Down~\cite{anderson2018bottom} regional features as image input, which are extracted by an object detector pre-trained in natural images and do not consider the visual domain-gap with paintings.

Finally, the text-summarization method LSA performs very poor in most of the metrics except in Skip-Thought, demonstrating that relying only on external knowledge without ``looking" into the image is not the best approach to generate a description for a painting. This is because with summarization methods: 1) it is not possible to control the generation process for the different topics, 2) they heavily rely on the accuracy of the knowledge retrieval module, and 3) descriptions cannot be generated when the external knowledge of a painting does not exist. According to a random subset of 150 annotated images,\footnote{Details can be found in Appendix D.} 80.7\% of the samples do not have a specific article on Wikipedia. However, 71.3\% of the them do have artist article, which is useful to find knowledge but not enough for producing a relevant painting description on its own. Moreover, even when retrieving a non-relevant article, our method can generate correct sentences for the \textcolor{colcontent}{\textbf{content}} and the \textcolor{colform}{\textbf{form}} topics. The ratio of slots per topic on the SemArt dataset, \ie, average number of slots in \textcolor{colcontent}{\textbf{content}}/\textcolor{colform}{\textbf{form}}/\textcolor{colcontext}{\textbf{context}} sentences, are 0.98/0.91/2.12, respectively, which shows that \textcolor{colcontent}{\textbf{content}} and \textcolor{colform}{\textbf{form}} require less external data than \textcolor{colcontext}{\textbf{context}}. Finally, the \textit{artistic attributes} predicted from the image are also used to fill the slots (Sec.~\ref{sec:knowledge_fill}), which may yield to meaningful information even when the retrieved knowledge fails.

\section{Conclusion}
We proposed the first multi-topic knowledgeable framework for art description generation. To generate multi-topic descriptions, we annotated an art description dataset with sentence-level topic labels. We explored this problem from multiple views, including proposing two types of topic decoder and experimenting with un/semi-supervised as well as supervised settings. Besides, we introduced the use of external knowledge to enhance the background information in the description of the painting. Comprehensive evaluation and comparison showed the effectiveness of our method, which we hope contributes to guide future research.

\vspace{-15pt}
\paragraph{Acknowledgement}
This work is partly supported by JSPS KAKENHI Grant Numbers JP20K19822 and JP18H03264, as well as ROIS NII Open Collaborative Research 2021-21S1002.

\appendix
\begin{appendices}

\section{Illustration of Different Decoders}
\begin{figure}
    \centering
    \includegraphics[width=0.99\linewidth]{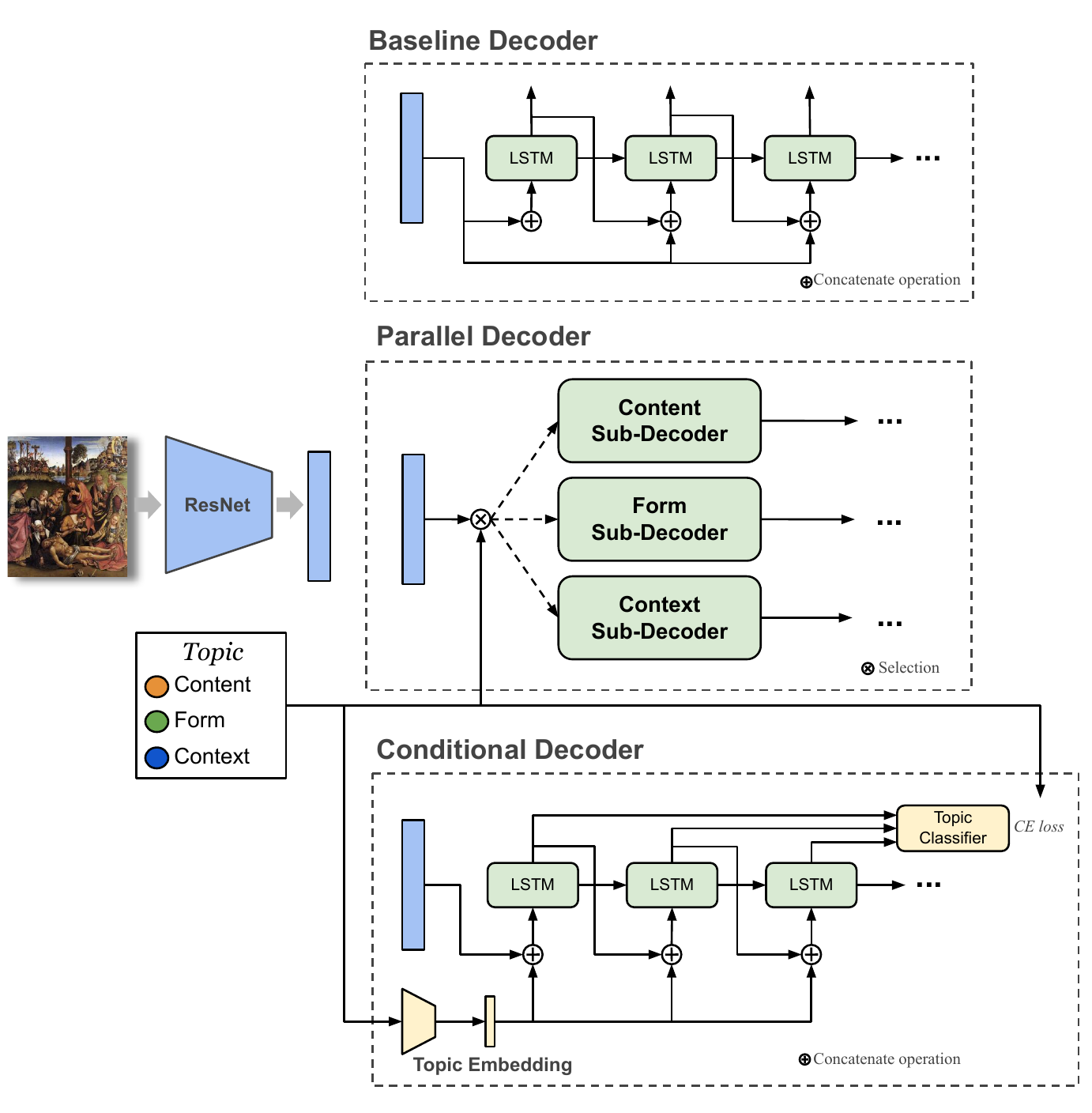}
    \caption{\textbf{Illustration of Different Decoders.} Baseline decoder and two variants of topic decoder.}
    \label{fig:topic_decoder}
\end{figure}

We present the details of different decoders in Fig.~\ref{fig:topic_decoder}. The formalization is provided in the main paper. In the baseline decoder, the context visual feature is calculated by weighting feature vectors from each region of the image and fed into the LSTM in each time step. For simplification, we omit the hidden state initialisation and the attention-based context visual feature calculation in the figure. In the parallel decoder, $N_\text{topic}$ baseline decoders are employed as sub-decoders. Apart from the visual feature, it uses the topic label as an additional input, which is responsible for selecting the corresponding decoder for the given topic. Finally, for the conditional decoder, the input topic label is first embedded into a topic embedding. Then, the topic embedding is concatenated to the visual feature and the previous hidden state for each time step. We omit the visual feature concatenation operation in the conditional decoder figure for simplification. After the whole decoding process, the generated masked sentence is fed to a topic classifier to ensure that the sentence belong to the correct topic.

\section{Implementation Details}
\label{sec:impl_detail}
We implement all our models with PyTorch \cite{paszke2019pytorch}. We optimize the topic decoder with the Adam \cite{kingma2014adam} with a learning rate of $5 \times 10^{-4}$, which decays at a rate of $0.8$ every $10$ epochs. The batch size is set to $32$. We extract $L=14 \times 14$ with $D=2,048$ feature maps from the layer before the last pooling layer of a pre-trained ResNet101 \cite{he2016deep}. For predicting artistic attributes, we use a four-branch attribute predictor model \cite{garcia2019context}. The dimensions of the LSTM-based decoder's hidden states and word embeddings are fixed to 512 for all of the models discussed herein. In the topic conditional decoder, the dimensionality of the topic embedding is set to 20. DrQA and BERT hyperparameters are set as in \cite{chen2017reading} and \cite{devlin2019bert}, respectively. At test time, we employ the beam search for generating text, where a beam size of 5 is empirically selected for all the topic decoder variants.

\section{Training Details}
\label{sec:training}
For the image encoder, we use a pre-trained ResNet \cite{he2016deep} that does not need to be trained. For the decoders, the baseline decoder is trained as the standard captioning model \cite{vinyals2015show}, where the whole description is used as ground truth caption for an image. While during training the topic decoder, the ground truth description for an image is split into $N_\text{topic}$ parts. Sentences with the same topic label are appended together as a topic-specific description. In the parallel decoder, each sub-decoder is trained independently with its topic-specific description. In training the conditional decoder, the topic-specific description are selected according to the topic label input to the decoder. In the topic classifier part, we employ the continuous approximation technique proposed by Hu \etal \cite{hu2017toward} to avoid sampling words from a probability distribution, so that the decoder and classifier can be trained in an end-to-end manner. Not all the comments contain the three topics. During training, if a comment does not span the \textit{e.g.,} \textit{form} topic, the \textit{form} decoder is not trained with that image.

In the knowledge retrieval part, both attributes prediction model and object detection model are pre-trained. While the DrQA \cite{chen2017reading} knowledge retriever adapts a non-machine-learning method. Thus, no optimization is needed in this part. In the knowledge extraction and filling part, BERT is trained with art descriptions. The input is a masked sentence and a list of candidate words, where the masks are generated by replacing the named-entities with their entity type, and candidate words are the named-entities that being replaced. The ground truth is the original sentence before masking. Note that to avoid trivial solutions, the candidate words are extracted from the whole paragraph of description while the input sentence is one short sentence.

\begin{table}[]
\small
\centering
\caption{\textbf{Knowledge retrieval.} Using attributes and objects words as query.}
\begin{tabularx}{0.48\textwidth}{l c c c c}
\toprule
Criterion & Num. Articles & Top-1 & Top-5 & Top-10 \\
\midrule
Correct articles & 29 & 0 & 3.4 & 3.4  \\
Theme articles & 3 & 1.5 & 1.5 & 1.5  \\
Author articles & 107 & 13.7 & 36.7 & 46.0  \\ \midrule
All articles & 150 & 13.8 & 36.6 & 45.5  \\
\bottomrule
\end{tabularx}
\label{tab:knwolwedge_words}
\end{table}

\begin{table}[]
\small
\centering
\caption{\textbf{Knowledge retrieval.} Using attributes and objects words, as well as generated masked sentences as query.}
\begin{tabularx}{0.48\textwidth}{l c c c c}
\toprule
Criterion & Num. Articles & Top-1 & Top-5 & Top-10 \\
\midrule
Correct articles & 29 & 0 & 0 & 3.4  \\
Theme articles & 3 & 0 & 0 & 0  \\
Author articles & 107 & 3.6 & 9.5 & 16.8  \\ \midrule
All articles & 150 & 5.0 & 10.5 & 17.5  \\
\bottomrule
\end{tabularx}
\label{tab:knwolwedge_words_template}
\end{table}

\begin{figure*}
    \centering
    \includegraphics[width=0.9\linewidth]{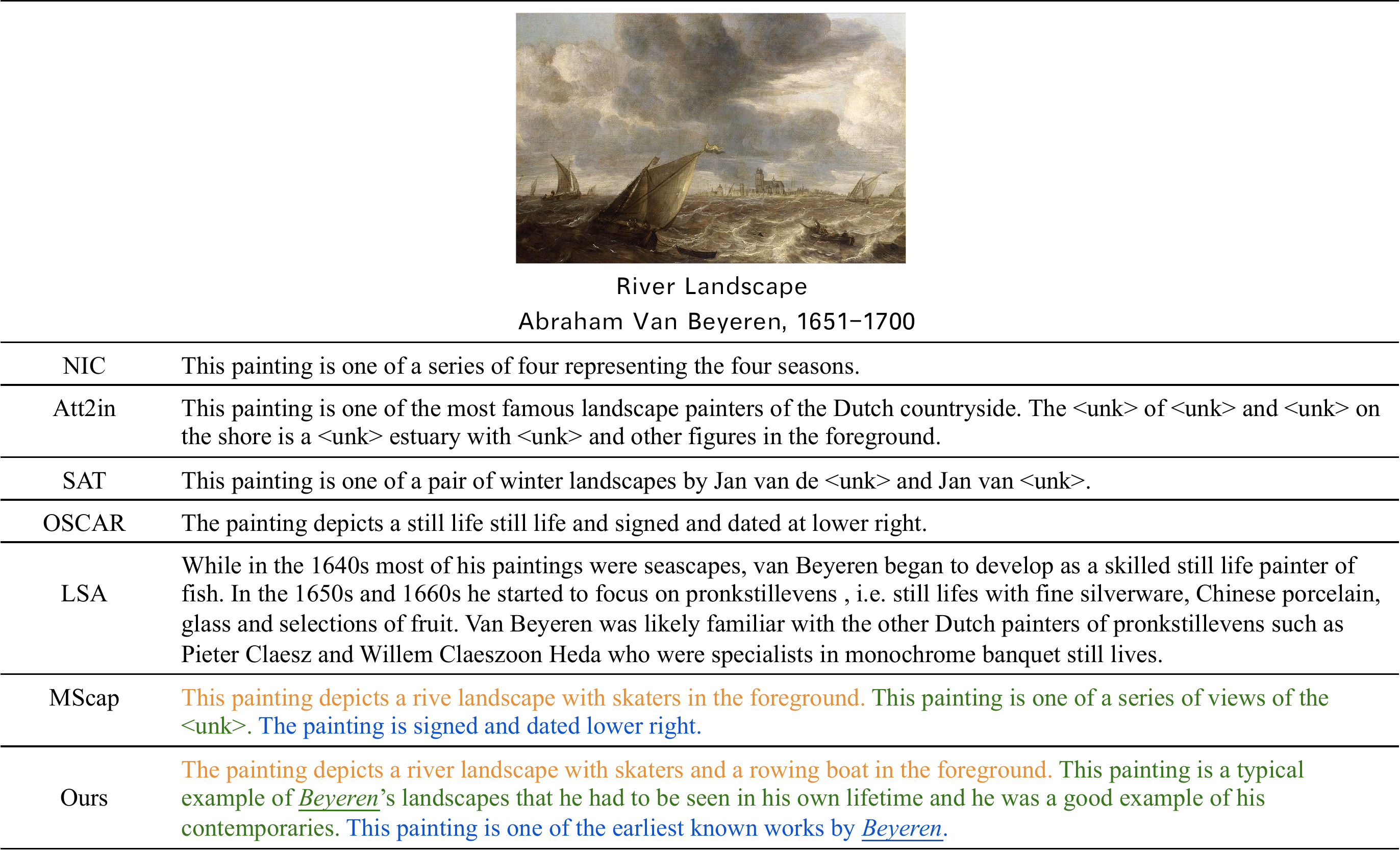}
    \caption{\textbf{Quantitative comparison with different methods.}}
    \label{fig:comp}
\end{figure*}

\begin{figure*}
    \centering
    \includegraphics[width=0.9\linewidth]{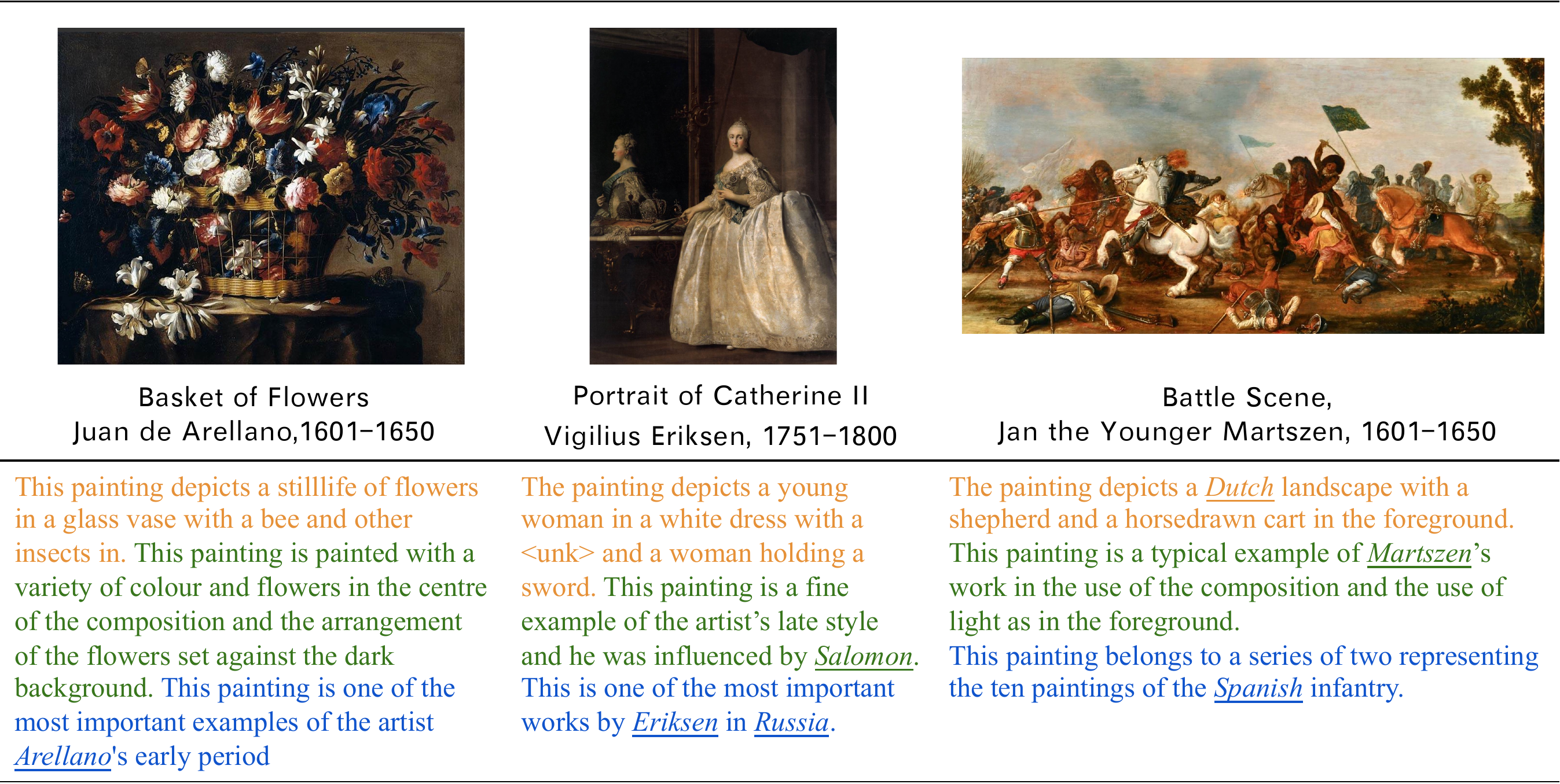}
    \caption{\textbf{More quantitative results produced by our framework.}}
    \label{fig:more_results}
\end{figure*}

\section{Knowledge Retrieval Module Evaluation}
For evaluating the knowledge retrieval module, we annotate a small number of paintings (150) with their correspondent Wikipedia article. Not all the images possess exact associated Wikipedia article. However, articles related to the painting's author or theme can also provide useful information. Considering these factors, we first prepare several candidate Wikipedia articles for each painting and annotate each article with one label out of the following five labels:

\begin{itemize}[noitemsep,topsep=0pt]
\item \textbf{Correct} the article is about the exact painting.

\item \textbf{Theme} the article is related to the content of the painting, \eg myth, person, event, concept, \etc.

\item \textbf{Author} the article is about the author. 

\item \textbf{Ambiguation} the article is about a painting with the same name but not the exact one, \ie. created by another author.

\item \textbf{Incorrect} unrelated article.
\end{itemize}

Among them, articles with Correct, Theme or Author labels are regarded as positive articles that can provide useful information, while Ambiguation and Incorrect correspond to negative articles. In total, we have annotated $450$ articles for $150$ paintings (3 articles for each painting).

We evaluate the accuracy of the knowledge retrieval module by comparing the sorted list of articles from our retriever with the annotated Wikipedia articles, and find the position in which the annotated article is returned. In this way, we measure recall at $k$ (R@$k$) metric with different values of $k$ (\eg, $k$ = {1, 5, 10}). R@$k$ represents the percentage of samples whose annotated article is returned within the top $k$ positions by our retriever. As we have different labels for the annotated articles, we calculate the metrics for the different type of articles.

Table~\ref{tab:knwolwedge_words} shows the evaluation results using attributes and objects words as query, as in the main paper. We can observe that the useful articles from our retriever mostly come from the author articles. We have also explored to incorporate the generated masked sentences into the query, whose results are shown in Table~\ref{tab:knwolwedge_words_template}. Comparing the two tables, we find that the incorporation of masked sentences has a negative impact in the knowledge retriever, as these sentences occupy a large proportion in the query but do not contain much specific information.

\section{More Qualitative Results}
Here we show the generated sentences by all the methods evaluated in the main paper and provide more qualitative results of our proposed method. Figure \ref{fig:comp} shows, the qualitative comparison of different methods in Section 4.2. In Figure \ref{fig:more_results}, three more examples of descriptions generated by our method are shown. 

\end{appendices}

{\small
\bibliographystyle{ieee_fullname}
\bibliography{egbib}
}

\end{document}